\documentclass[10pt,twocolumn,letterpaper]{article}

\usepackage{iccv}
\usepackage{times}
\usepackage{epsfig}
\usepackage{graphicx}
\usepackage{amsmath}
\usepackage{amssymb}
\usepackage{multirow}
\usepackage{pifont}
\usepackage{arydshln}
\usepackage{booktabs}
\usepackage{bbding}
\usepackage[linesnumbered,lined,boxed,commentsnumbered]{algorithm2e}
\usepackage{caption}
\usepackage[dvipsnames]{xcolor}
\usepackage[percent]{overpic}

\usepackage[accsupp]{axessibility} 

\usepackage{color}
\usepackage{amsmath}

\usepackage{colortbl}  
\usepackage{xcolor}
\usepackage{array}

\definecolor{hollywoodcerise}{rgb}{0.96, 0.0, 0.63}
\definecolor{lasallegreen}{rgb}{0.03, 0.47, 0.19}
\definecolor{hanpurple}{rgb}{0.32, 0.09, 0.98}
\definecolor{green(pigment)}{rgb}{0.0, 0.65, 0.31}

\usepackage[breaklinks=true,bookmarks=false]{hyperref}
\hypersetup{colorlinks,linkcolor={red},citecolor={hollywoodcerise},urlcolor={red}}  

\iccvfinalcopy 

\ificcvfinal\fi

\begin{document}
\title{OmniZoomer: Learning to Move and Zoom in on Sphere at High-Resolution}

\author{Zidong Cao$^{1}$ \quad Hao Ai$^{1,2}$\thanks{
Intern at ARC Lab, Tencent PCG.
} \quad Yan-Pei Cao$^{2}$ \quad Ying Shan$^{2}$ \quad Xiaohu Qie$^{2}$ \quad Lin Wang$^{1}$$^{,3}$\thanks{Corresponding author (e-mail: linwang@ust.hk)}\\
$^{1}$ AI Thrust, HKUST(GZ) \quad $^{2}$ARC Lab, Tencent PCG \quad $^{3}$Dept. of CSE, HKUST \\
{\tt\small caozidong1996@gmail.com,  hai033@connect.hkust-gz.edu.cn, caoyanpei@gmail.com}\\{\tt\small yingsshan@tencent.com, tigerqie@tencent.com, linwang@ust.hk}}


\twocolumn[{
\renewcommand\twocolumn[1][]{#1}%
\maketitle
\begin{center}
\centering
\vspace{-20pt}


\includegraphics[width=.95\textwidth]{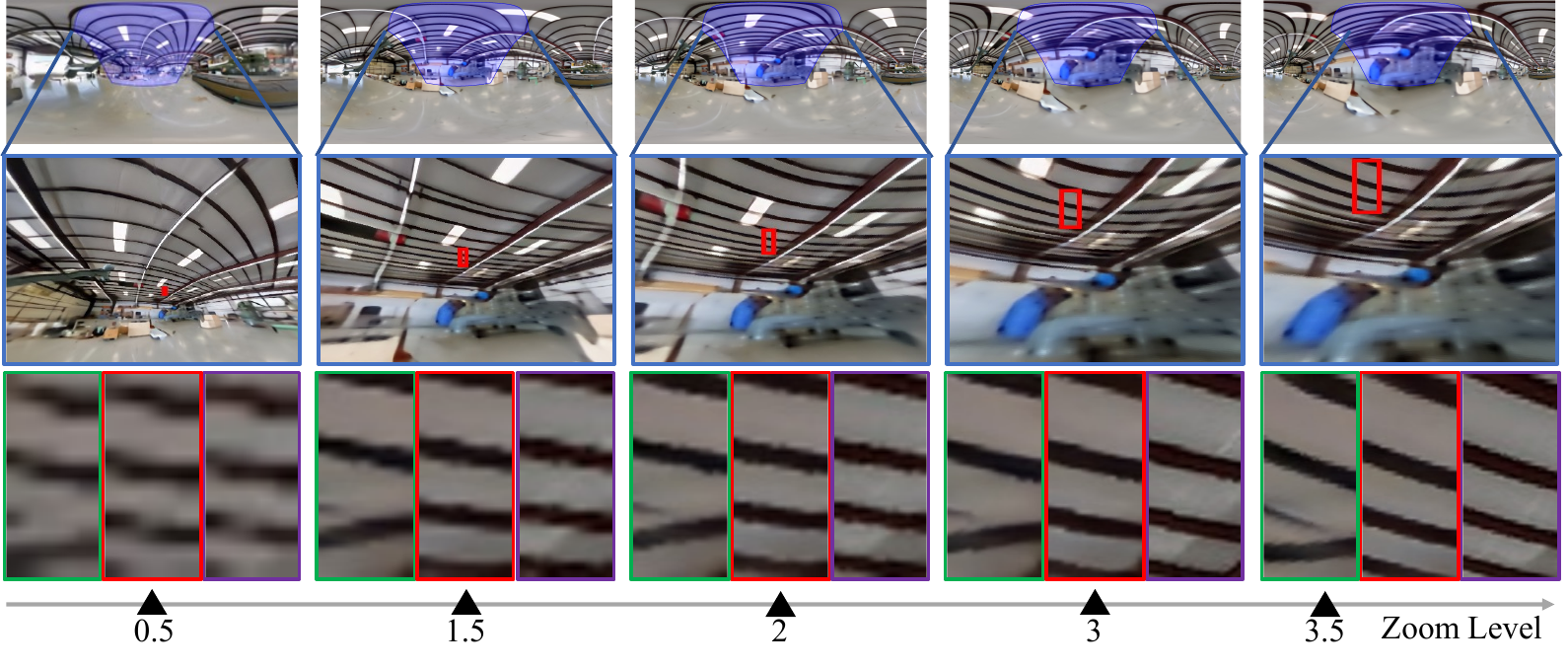}
\captionsetup{font=small}
\end{center}
\vspace{-20pt}
\captionof{figure}{\textbf{From top to bottom}: 
equirectangular projection (ERP) of omnidirectional images, perspective projection from ERP images with a specific field-of-view (FoV), \wrt the \textcolor{blue}{blue} regions, and results from \textcolor{OliveGreen}{LAU-Net}, \textcolor{Red}{OmniZoomer} and \textcolor{Purple}{ground truth}. \textbf{From left to right}: increasing zoom levels. Our approach can move to the object of interest, and freely zoom in and zoom out on omnidirectional images, which can recover clear and preserved textural details with increasing zoom levels.}

\vspace{15pt}
\label{fig:1}

}]
\let\thefootnote\relax\footnote{$^{*}$ Intern at ARC Lab, Tencent PCG.

\hspace*{0.16cm} $^{\dag}$ Corresponding author}
\ificcvfinal\thispagestyle{empty}\fi

\begin{abstract}
Omnidirectional images (ODIs) have become increasingly popular, as their large field-of-view (FoV) can offer viewers the chance to freely choose the view directions in immersive environments such as virtual reality. The M\"obius transformation is typically employed to further provide the opportunity for movement and zoom on ODIs, but applying it to the image level often results in blurry effect and aliasing problem. In this paper, we propose a novel deep learning-based approach, called \textbf{OmniZoomer}, to incorporate the M\"obius transformation into the network for movement and zoom on ODIs. By learning various transformed feature maps under different conditions, the network is enhanced to handle the increasing edge curvatures,
which alleviates the blurry effect. Moreover, to address the aliasing problem, 
we propose two key components.
Firstly, to compensate for the lack of pixels for describing curves, we enhance the feature maps in the high-resolution (HR) space and calculate the transformed index map with a spatial index generation module.
Secondly, considering that ODIs are inherently represented in the spherical space, we propose a spherical resampling module that combines the index map and HR feature maps to transform the feature maps for better spherical correlation. The transformed feature maps are decoded to output a zoomed ODI. Experiments show that our method can produce HR and high-quality ODIs with the flexibility to move and zoom in to the object of interest. Project page is available at \url{http://vlislab22.github.io/OmniZoomer/}.

\end{abstract}
\vspace{-20pt}

\section{Introduction}
\label{sec:intro}
Omnidirectional images (ODIs) have garnered significant attention as a means to maximize the amount of content and context captured within a single image, and there is a growing demand for utilizing such visual content within devices, \eg, mobile apps and head-mounted displays (HMDs) for virtual reality (VR)~\cite{tateno2018distortion}. To provide an interactive experience, these devices enable users to control the view direction. However, most $360^{\circ}$ cameras have a fixed focal length and do not support optical zoom, which causes the apparent size of objects in ODIs fixed. This limits the immersive experience when users expect to move and zoom in to an object of interest to see more details.

\begin{figure}[t]
    \centering
    \captionsetup{font=small}
    \includegraphics[width=1\linewidth]{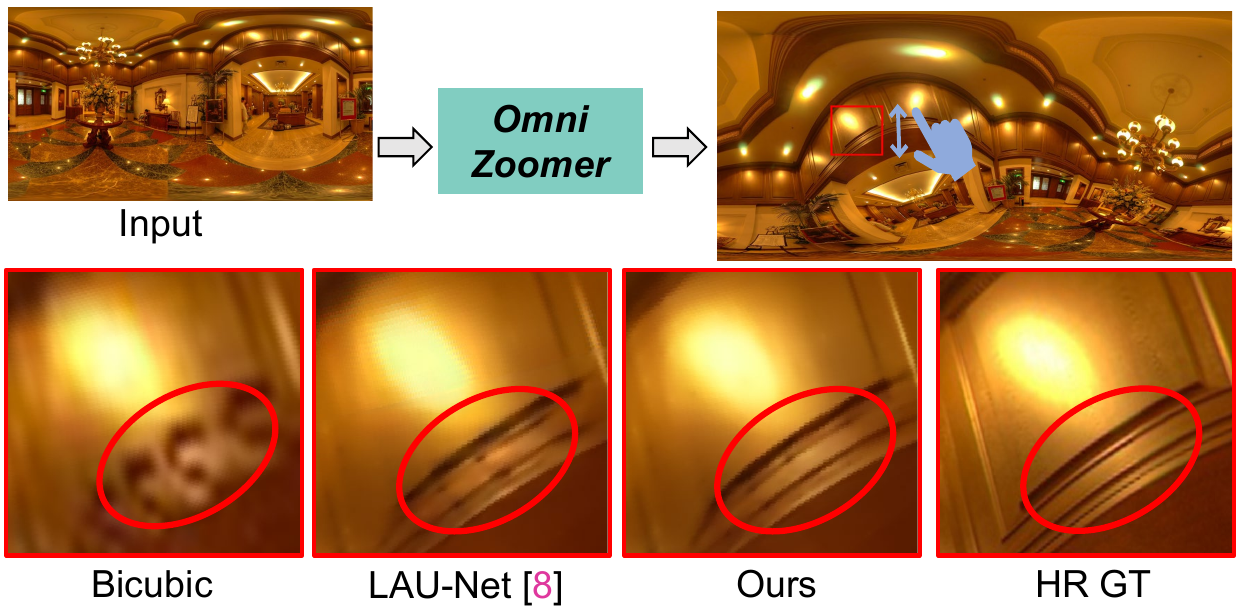}
\caption{Visual comparisons of different methods for movement and zoom. Our OmniZoomer predicts more continuous lines.}
\label{fig:2}
\end{figure}

Generally, there exist three solutions to zoom in on the equirectangular projection (ERP) format ODIs or their perspective patches. The first is to zoom in on ERP images uniformly. However, as ERP images have non-uniform pixel density in different latitudes~\cite{deng2021lau}, uniform zoom can severely distort the object shapes. The second is to zoom in on perspective patches projected from ODIs. As the perspective patches of ODI have uniform pixel density~\cite{Eder2019TangentIF}, distortion problem can be solved. However, due to the limited FoV, these patches only concentrate on local regions and ignore the relationship between each other during transformations. 
Thirdly, M\"obius transformation has recently been employed to provide movement and zoom freedom on ODIs~\cite{schleimer2016squares, ferreira2017local, mizuguchi2018basic}. It is the only conformal bijective transformation on the sphere that preserves angles. However, applying M\"obius transformation on the image level often leads to blurry and aliasing problems due to two reasons.
Firstly, zoom-in makes a portion of the ODIs enlarged, making the enlarged region blurry and pixelated. Moreover, if $360^{\circ}$ cameras are placed vertically, the ODIs suffer from distortion mainly in high-latitude regions and remain lots of straight lines in equator regions. After transformations, the appearance of the vertically captured ODIs varies greatly, resulting in more curves in both high-latitude and equator regions~(See Fig.~\ref{fig:2}). Describing these curves with the same amount of pixels that originally represent straight lines becomes challenging. 

To obtain high-quality ODIs after movement and zoom, in this paper, we propose a novel deep learning-based approach, dubbed \textbf{OmniZoomer}, to incorporate the M\"obius transformation into the network for freely moving and zooming in on ODIs, as shown in Fig.~\ref{fig:1}. By learning transformed feature maps in various conditions, the network is enhanced to handle the increasing curves caused by movement and zoom, as well as the inherent spherical distortion in ODIs. In this case, the blurry effect can be solved to some extent, but the aliasing problem still exists, such as edge discontinuity and shape distortion (See Fig.~\ref{fig:ab-order}(c)).

To further address the aliasing problem, we propose two key components. Firstly, to compensate for the lack of pixels for describing curves, we propose to enhance the extracted feature maps to high-resolution (HR) space before the transformation. The HR feature maps contain more fine-grained textural details, and are sufficient to represent the increasing curvatures and maintain the object shapes precisely. We then propose a spatial index generation module (Sec.~\ref{sec:grid}) to calculate the transformed index map based on the HR feature maps and M\"obius transformation matrix, which can be conducted on the HR feature space.
Although applying M\"obius transformation on HR images with existing super-resolution (SR) methods~\cite{lim2017enhanced, wang2018esrgan, zhang2018image} can serve the same purpose,
this solution is sub-optimal because the models might not handle the increasing curves (See Tab.~\ref{tab:msr} and Fig.~\ref{fig:comparex8}).
In addition, some image warping methods~\cite{son2021srwarp, lee2022learning} can learn the warping process in the network but are constrained to estimate spatial-varying grids on the 2D plane, rather than the sphere. There are also some SR models designed for ODIs~\cite{deng2021lau, yoon2022spheresr}. However, they are limited to vertically captured ODIs or predetermined data structures. 

\begin{figure*}[t]
    \centering
\includegraphics[width=.98\linewidth]{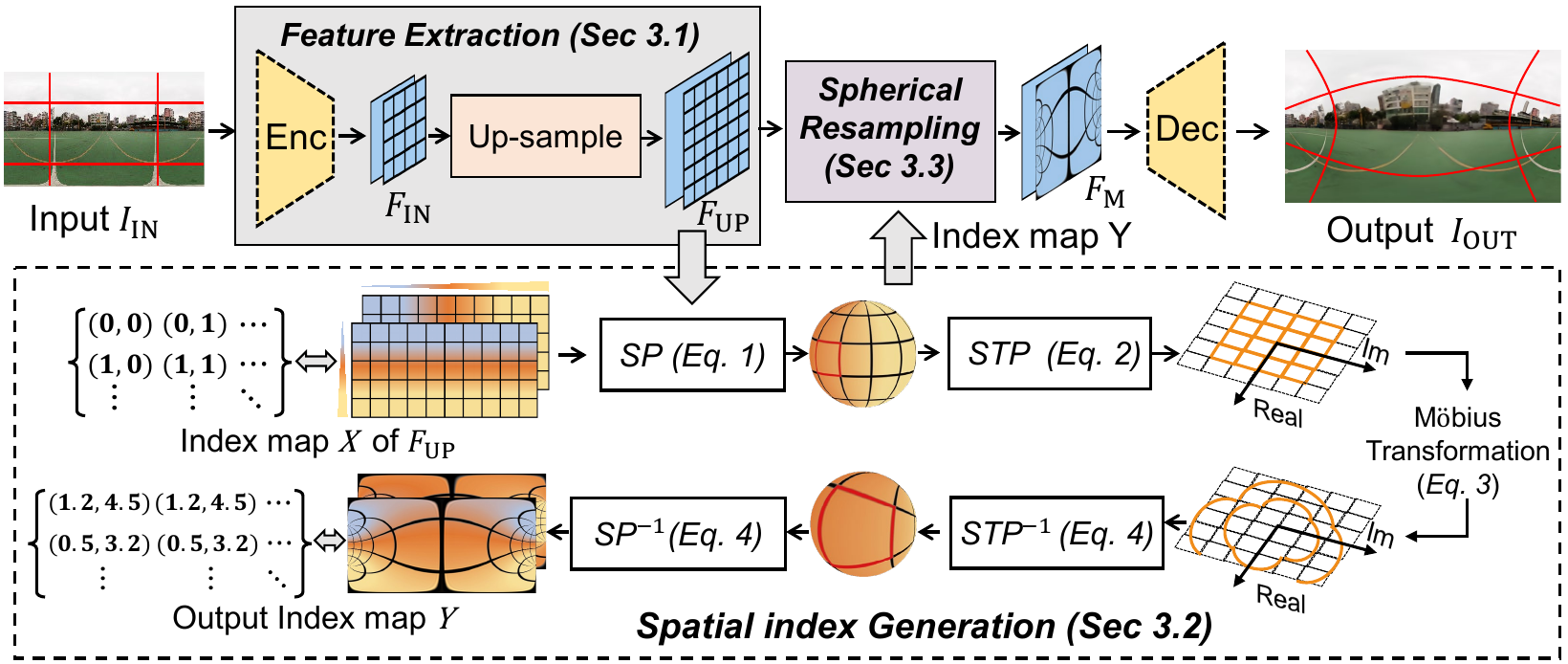}
\captionsetup{font=small}
\caption{\textbf{The overall pipeline of the proposed OmniZoomer.} With the spatial index generation module and spherical resampling module, OmniZoomer can provide users with a flexible way to zoom in and out to objects of interest, such as the enlarged center building.}
\label{fig:4}
\end{figure*}

Subsequently, we propose a spherical resampling module that combines the HR feature maps and transformed index maps for feature map transformation. The spherical resampling is inspired by the inherent spherical representation of the ODIs and the spherical conformality of M\"obius transformation. It resamples based on the spherical geodesic of two points on the sphere, which better relates the original HR feature maps and transformed ones. With HR feature representation and the spherical resampling module, OmniZoomer alleviates the blurry effect and aliasing problem substantially, enabling moving and zooming in to an object of interest on ODIs with preserved shapes and continuous curves. Finally, these feature maps are processed with a decoder to output a zoomed ODI. After movement and zoom, OmniZoomer can generate more precise visual results with clear textural and structural details (See Fig.~\ref{fig:2}). 

As collecting real-world ODI pairs under M\"obius transformation is difficult, we propose a dataset based on ODI-SR dataset~\cite{deng2021lau}, dubbed ODIM dataset, containing synthesized ODIs with various M\"obius transformations. We evaluate the effectiveness of OmniZoomer on the ODIM dataset under various M\"obius transformations and up-sampling factors. The experimental results show that OmniZoomer outperforms existing methods quantitatively and qualitatively.

The main contributions of this paper can be summarized as follows: (\textbf{I}) We propose a novel deep learning-based approach, called \textit{OmniZoomer}, to incorporate the M\"obius transformation into the deep network. (\textbf{II}) We enhance the feature maps to HR space and calculate the HR index map with a spatial index generation module. We also propose a spherical resampling module for better spherical correlation. (\textbf{III}) We establish ODIM dataset for supervised training. Compared with existing methods, OmniZoomer achieves the state-of-the-art performance under various M\"obius transformations and up-sampling factors.

\section{Related Work}
\label{sec:related}

\noindent \textbf{Application of M\"obius Transformation.} One of the main immersive experience in $360^{\circ}$ devices is the control by the viewers. Although current devices can provide the opportunity to control view directions and field-of-views (FoVs)~\cite{baur2016exploring}, the zoom quality needs to be improved to see more details~\cite{Potengy2018deMP}.
M\"obius transformation has been applied on ODIs, including straight line rectification~\cite{penaranda2018real, ferreira2017local, ferreira2020bounded}, , stereo pairs rectification~\cite{geyer2003conformal}, and rotation and zoom~\cite{schleimer2016squares}. However, these methods operate on the image level, whose performance relies heavily on the quality of the raw ODI. 

Recently, in ~\cite{wu2022view}, M\"obius transformation has been employed in deep learning for feature augmentation. Especially, ~\cite{wu2022view} fuses the features, which are applied multiple transformations to predict the raw ODI and address the spherical distortion. However, ~\cite{wu2022view} dose not generate the HR and high-quality transformed ODIs. Moreover, M\"obius transformation has also been applied for data augmentation~\cite{zhou2021data}, activation function~\cite{mandic2009complex, ozdemir2011complex}, pose estimation~\cite{azizi20223d}, and convolutions~\cite{mitchel2022mobius}.
Although M\"obius convolution~\cite{mitchel2022mobius} shows a strong capability of spherical equivalence, it requires the spherical harmonic transform in each convolution block, resulting in low computational efficiency. 
\textit{In this work, we propose a learning-based approach to improve the textural and structural details of ODIs when moving and zooming in to an object of interest.}

\noindent\textbf{Image Warping.}  It is widely utilized in various tasks, \eg, optical flow estimation~\cite{brox2004high} and video SR~\cite{liang2022vrt}. Generally, it is conducted by calculating transformed spatial indices, and resampling information from the input images based on the transformed indices~\cite{jaderberg2015spatial}. Considering jagging and blurry effects of image warping~\cite{son2021srwarp}, SRWarp interprets the image warping as a spatially-varying SR problem and proposes an adaptive warping layer to estimate the rotation during warping. SRWarp also shows that simply concatenating existing SR models~\cite{lim2017enhanced, wang2018esrgan, zhang2018image, hu2019meta} with warping operation is sub-optimal. Furthermore, LTEW~\cite{lee2022learning} estimates the varied shape and integrates the priors into an implicit representation in the Fourier space. \textit{Differently, we focus on transforming and resampling on the sphere with a curved surface. Our proposed spherical resampling module outperforms these warping methods significantly. (See Tab.~\ref{table:ab-interpolation})}.

\noindent \textbf{ODI Super-Resolution.}
Traditional ODI SR methods primarily utilize a sequence of low-resolution (LR) ODIs to stitch an HR ODI \cite{arican20091, arican2011joint, bagnato2010plenoptic, kawasaki2006super, nagahara2000super}. Recently, \cite{fakour2018360} and \cite{nishiyama2021360} propose learning-based SR methods that incorporate the distortion maps to tackle spherical distortions. \cite{ozcinar2019super} employs adversarial learning for ODI SR, but only treats ODIs as 2D planar images. 
Observing that different latitudes have non-uniform pixel densities, LAU-Net~\cite{deng2021lau} crops ODIs into different latitude bands and dynamically up-samples these bands. However, after transformations, \eg, movement and zoom, the bands can not be simply cropped along latitudes. SphereSR~\cite{yoon2022spheresr} proposes to super-resolve an LR ODI to an HR ODI with arbitrary projection types. Nevertheless, the predetermined spherical data structure can not adapt to the transformed ODIs. \textit{Unlike these methods, we address a new task of incorporating the M\"obius transformation into the network to move and zoom in to the object of interest on ODIs with high-quality textural and structural details.}
\begin{figure}[t!]
    \captionsetup{font=small}\centering\includegraphics[width=1\linewidth]{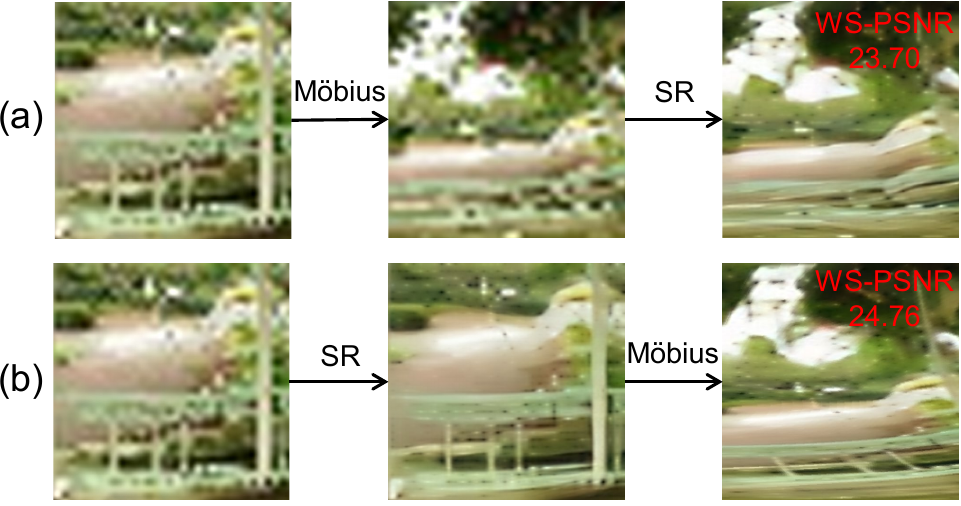}
\caption{Comparisons of directly applying M\"obius transformation on the ODI and on the super-resolved ODI.}
\vspace{-15pt}
\label{fig:6}
\end{figure}

\section{Methodology}
\label{method}

\noindent{\textbf{Overview.}} As shown in Fig.~\ref{fig:4}, we propose a novel end-to-end pipeline, dubbed \textit{OmniZoomer}, which allows for free movement of the ``eyes" to objects of interest and zooming in directly on the sphere with preserved shapes and high-quality textural details. Firstly, we extract HR feature maps $F_{\text{UP}} \in \mathbb{R}^{H\times W \times C}$ from the input ODI $I_{\text{IN}} \in \mathbb{R}^{h\times w \times 3}$ through an encoder and an up-sampling block (Sec.~\ref{sec:upsample}). With $F_{\text{UP}}$'s index map $X \in \mathbb{R}^{H\times W \times 2}$ as the input, we propose the spatial index generation module (Sec.~\ref{sec:grid}) to apply the M\"obius transformation~\cite{Kato2015MobiusTA} with arbitrary parameters on $X$ for the transformed spatial index map $Y \in \mathbb{R}^{H\times W \times 2}$.
Note that the channel numbers of $X$ and $Y$ indicate the longitude and latitude, respectively. Subsequently, we introduce a spherical resampling module (Sec.~\ref{sec:Slerp}) that generates the transformed HR feature maps $F_{\text{M}} \in \mathbb{R}^{H\times W \times C}$ by resampling the pixels on the sphere guided by the transformed index map $Y$. Finally, we decode the feature maps to output a zoomed-in ODI where the region of interest is a clear close-up shot. The decoder consists of three ResBlocks~\cite{lim2017enhanced} and a convolution layer.
We take the same parameters used in the spatial index generation module to transform the HR ground truth ODIs, and employ the $L1$ loss as the supervision loss. We now provide detailed descriptions of these components. 
\subsection{Feature Extraction}
\label{sec:upsample}

Given an ODI $I_{\text{IN}}\in\mathbb{R}^{h\times w \times 3}$ with the ERP format, we first apply an encoder consisting of several convolution layers to extract the feature maps $F_{\text{IN}}\in\mathbb{R}^{h\times w \times C}$. Accordingly, we design an upsampling block with several pixel-shuffle layers~\cite{Shi2016RealTimeSI} to generate the HR feature maps $F_{\text{UP}}\in\mathbb{R}^{H\times W \times C}$, where $H=s * h$, $W=s * w$, $s$ is the scale factor and $C$ is the channel number. Especially, we apply the M\"obius transformation on the HR feature maps based on two considerations: 1) \textit{ The blurry effect on image level}. By learning various transformations, the extracted feature maps demonstrate an enhanced representation capability in handling increasing edge curvatures and solving the blurry effect. 2) \textit{The aliasing problem.} For instance, in Fig.~\ref{fig:6}(a), due to the insufficient pixels to describe continuous and clear curves after transformations, the shape of the railing is distorted. 
Moreover, the aliasing problem is challenging to tackle even if super-resolving the transformed ODIs. 
In contrast, applying the M\"obius transformation to the super-resolved ODI has a significant improvement, as demonstrated in Fig.~\ref{fig:6}(b).

\subsection{Spatial Index Generation }
\label{sec:grid}

In this section, we apply the M\"obius transformation on the spatial index map $X$ of HR feature maps $F_{\text{UP}}$ and generate the transformed spatial index map $Y$ for the subsequent resampling operation. M\"obius transformation is known as the only conformal bijective transformation between the sphere and the complex plane. To apply the M\"obius transformation on the HR feature maps $F_{\text{UP}}$, we first use spherical projection (SP) to project the spatial index map $X$ from spherical coordinates $(\theta, \phi)$ (where $\theta$ represents the longitude and $\phi$ represents the latitude) to the Riemann sphere ${\mathbb {S}^2 = \{(x,y,z) \in \mathbb {C}^3 | x^2+y^2+z^2=1}\}$, formulated as:
\begin{equation}
\small
\text{SP}: \begin{pmatrix}
     x  \\
     y \\
     z \\
\end{pmatrix} = 
\begin{pmatrix}
     \cos(\phi)\cos(\theta) \\
     \cos(\phi)\sin(\theta) \\
     \sin(\phi) \\
\end{pmatrix} .
\label{eq:2}
\end{equation}
Then, using stereographic projection (STP)~\cite{Eybpoosh2021ApplyingIS}, we can project a point $(x, y, z)$ of the Riemann sphere $\mathbb {S}^2$ onto the complex plane and obtain the projected point ($x'$, $y'$). Let point $(0,0,1)$ be the pole, STP can be formulated as:
{\begin{equation}
\text{STP}: x' = {\frac {x }{1 - z}}\ ,\ y' = {\frac {y}{1 - z}}.
\label{eq:3}
\end{equation}}

Subsequently, given the projected point $p$ ($Z_p$ = $x'$+$i y'$) on the complex plane, we can conduct the M\"obius transformation with the following formulation:
{\begin{equation}
f(Z_p)={\frac {aZ_p+b}{cZ_p+d}}, 
\label{eq:1}
\end{equation}}
where $a$, $b$, $c$, and $d$ are complex numbers satisfying $ad-bc \neq 0$. Finally, we apply the inverse stereographic projection $\text{STP}^{-1}$ and inverse spherical projection $\text{SP}^{-1}$ to re-project the complex plane into the ERP plane:
\begin{equation}
\begin{split}
    \small
\renewcommand{\arraystretch}{1.5}
\setlength{\arraycolsep}{1.3pt}
    \text{STP}^{-1}: \begin{pmatrix}
         x  \\
         y \\
         z \\
    \end{pmatrix} &= 
    \begin{pmatrix}
         \frac{2x'}{1+x'^2+y'^2} \\
         \frac{2y'}{1+x'^2+y'^2} \\
         \frac{-1+x'^2+y'^2}{1+x'^2+y'^2} \\
        \end{pmatrix} \ ; \\
   \ \text{SP}^{-1}: \begin{pmatrix}
         \theta \\
         \phi \\
    \end{pmatrix} &= 
    \begin{pmatrix}
         \arctan(y/x) \\
         \arcsin(z) \\
        \end{pmatrix} \ .
\end{split}
    \label{eq:5}
\end{equation}

In summary, as shown in Fig.~\ref{fig:4}, we first project the input index map $X$ to the complex plane using SP (Eq.~\ref{eq:2}) and STP (Eq.~\ref{eq:3}), and then conduct the M\"obius transformation with Eq.~\ref{eq:1}, and generate the transformed index map $Y$ through the inverse STP (Eq.~\ref{eq:5}) and inverse SP (Eq.~\ref{eq:5}). After transformation, both the indices (represented with matrix and gradient color) and grid shapes (represented with black lines) in $Y$ have a noticeable change compared with $X$.

\begin{figure}[t]
    \centering
    \includegraphics[width=0.94\linewidth]{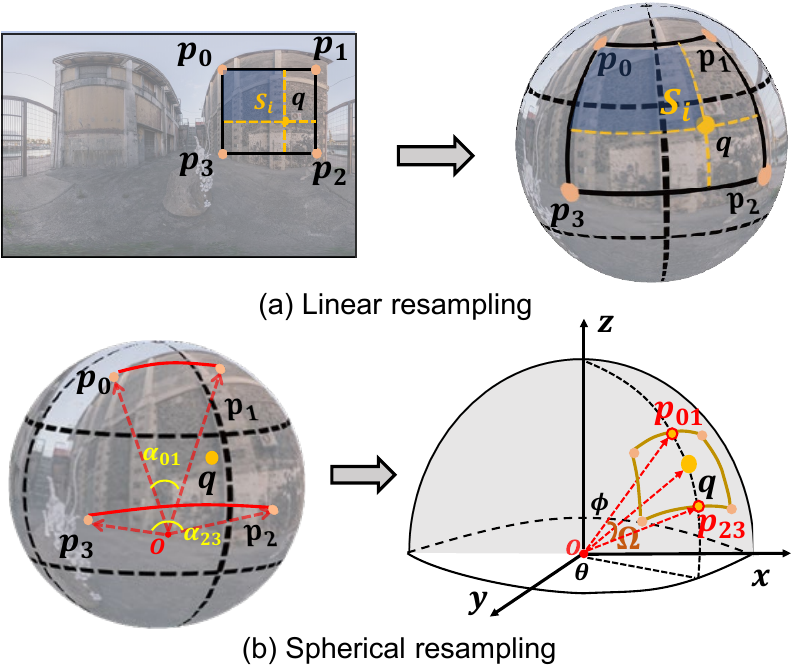}
    \captionsetup{font=small}
\caption{(a) Linear resampling is related to the the partial area $S_{i}$ diagonally opposite to the corner pixel $i$. (b) Spherical resampling considers the angles (\ie, $\alpha_{01}$, $\alpha_{23}$, $\Omega$) between points on the sphere, which are corresponding to the red solid curves.}
\vspace{-15pt}
\label{fig:7}
\end{figure}

\subsection{Spherical Resampling}
\label{sec:Slerp}
As spatial indices recorded in $Y$ are not equidistant, it is necessary to design a resampling method to calculate the feature values for the transformed feature maps $F_\text{M}$ based on $Y$. 
Generally, the resampling process can be divided into three steps. The first step is to determine the neighboring pixel set $N_{q}$ of the query pixel $q$, \ie, the four corner pixels $\{p_i \in N_q, i=0,1,2,3\}$, as illustrated in Fig.~\ref{fig:7}(a). The second step is to calculate the weight ${w_{q,p_i}}$ for each neighboring pixel $p_i \in N_{q}$, \ie, the partial area ${S_{i}}$. The third step is to calculate the weighted average of neighboring feature values: $F(q) = \sum_{p_i \in N_{q}}{w_{q, p_i}F(i)}$. 

Previous image warping methods, \eg, SRWarp~\cite{son2021srwarp} and LTEW~\cite{lee2022learning}, consider the rotation of the local varied grid with Jacobian matrix. In this case, the resampling bases are recalculated according to the grid rotation, and the resampling weight $w_{q, p_i}$ is re-projected to the new bases, either explicitly~\cite{son2021srwarp} or implicitly~\cite{lee2022learning}. 
Although these methods can be used directly for M\"obius transformation on ODIs, they can not deal with the spherical representation of ODIs due to two key reasons: 1) As shown in Fig.~\ref{fig:7}(a), the partial area (marked with blue region) in 2D plane is stretched non-uniformly when projected to the spherical surface due to spherical distortion; 
2) Although the resampling bases can be corrected with rotation, the resampling process is still limited to 2D plane, which is sub-optimal for describing the relationship between two points on the sphere.

\begin{table*}[t!]
\scalebox{0.82}{
\begin{tabular}{c||cccc||cccc}
\toprule
Scale & \multicolumn{4}{c||}{$\times 8$} & \multicolumn{4}{c}{$\times 16$} \\ \hline
\multirow{2}{*}{Method} & \multicolumn{2}{c|}{ODI-SR} & \multicolumn{2}{c||}{SUN 360} & \multicolumn{2}{c|}{ODI-SR} & \multicolumn{2}{c}{SUN 360} \\ \cline{2-9} & \multicolumn{1}{c|}{WS-PSNR} & \multicolumn{1}{c|}{WS-SSIM} & \multicolumn{1}{c|}{WS-PSNR} & WS-SSIM & \multicolumn{1}{c|}{WS-PSNR} & \multicolumn{1}{c|}{WS-SSIM} & \multicolumn{1}{c|}{WS-PSNR} & WS-SSIM \\ \hline \hline
Bicubic  & \multicolumn{1}{c|}{26.77} & \multicolumn{1}{c|}{0.7725} & \multicolumn{1}{c|}{25.87} & \multicolumn{1}{c||}{0.7103} & \multicolumn{1}{c|}{24.79} & \multicolumn{1}{c|}{0.7404} & \multicolumn{1}{c|}{23.87} & \multicolumn{1}{c}{0.6802} \\ \hline
EDSR-baseline${\rm (+Transform)}$~\cite{lim2017enhanced} & \multicolumn{1}{c|}{27.42} & \multicolumn{1}{c|}{0.7930} & \multicolumn{1}{c|}{26.97} & \multicolumn{1}{c||}{0.7468} & \multicolumn{1}{c|}{25.39}        & \multicolumn{1}{c|}{0.7572}        & \multicolumn{1}{c|}{24.66} & \multicolumn{1}{c}{0.7011} \\ \hline 
Ours-EDSR-baseline  & \multicolumn{1}{c|}{27.48} & \multicolumn{1}{c|}{0.7949}& \multicolumn{1}{c|}{27.15} & \multicolumn{1}{c||}{0.7526} & \multicolumn{1}{c|}{25.47} & \multicolumn{1}{c|}{\textbf{0.7600}}  & \multicolumn{1}{c|}{24.79} & \multicolumn{1}{c}{\textbf{0.7050}} \\ \hline
\hline
RRDB${\rm (+Transform)}$~\cite{wang2018esrgan}  & \multicolumn{1}{c|}{27.45} & \multicolumn{1}{c|}{0.7946} & \multicolumn{1}{c|}{27.10} & \multicolumn{1}{c||}{0.7515} & \multicolumn{1}{c|}{25.42} & \multicolumn{1}{c|}{0.7578} & \multicolumn{1}{c|}{24.72}  & \multicolumn{1}{c}{0.7033}  \\ \hline
RCAN${\rm (+Transform)}$~\cite{zhang2018image}  & \multicolumn{1}{c|}{27.46} & \multicolumn{1}{c|}{0.7906} & \multicolumn{1}{c|}{27.04} & \multicolumn{1}{c||}{0.7443} & \multicolumn{1}{c|}{25.45} & \multicolumn{1}{c|}{0.7541} & \multicolumn{1}{c|}{24.70}  & \multicolumn{1}{c}{0.7001}  \\ \hline
ETDS${\rm (+Transform)}$~\cite{chao2023equivalent} & \multicolumn{1}{c|}{27.38} & \multicolumn{1}{c|}{0.7912} & \multicolumn{1}{c|}{26.84} & \multicolumn{1}{c||}{0.7418} & \multicolumn{1}{c|}{25.42} & \multicolumn{1}{c|}{0.7572} & \multicolumn{1}{c|}{24.65} & \multicolumn{1}{c}{0.7012}\\ \hline
Omni-SR${\rm (+Transform)}$~\cite{wang2023omni} & \multicolumn{1}{c|}{27.45} & \multicolumn{1}{c|}{0.7920} & \multicolumn{1}{c|}{26.99} &
\multicolumn{1}{c||}{0.7463} & \multicolumn{1}{c|}{25.45} & 
\multicolumn{1}{c|}{0.7574} & \multicolumn{1}{c|}{24.68} &
\multicolumn{1}{c}{0.7010}\\ \hline 
LAU-Net${\rm (+Transform)}$~\cite{deng2021lau} & \multicolumn{1}{c|}{27.25} & \multicolumn{1}{c|}{0.7813}       & \multicolumn{1}{c|}{26.77}     & \multicolumn{1}{c||}{0.7363} & \multicolumn{1}{c|}{25.23} & \multicolumn{1}{c|}{0.7455} & \multicolumn{1}{c|}{24.49} & \multicolumn{1}{c}{0.6921} \\ \hline
SRWarp~\cite{son2021srwarp} & \multicolumn{1}{c|}{27.43} & \multicolumn{1}{c|}{0.7911} & \multicolumn{1}{c|}{27.12}  & \multicolumn{1}{c||}{0.7495} & \multicolumn{1}{c|}{25.40} & \multicolumn{1}{c|}{0.7570} & \multicolumn{1}{c|}{24.73} & \multicolumn{1}{c}{0.7014} \\
\hline
LTEW~\cite{lee2022learning}  & \multicolumn{1}{c|}{27.32} & \multicolumn{1}{c|}{0.7899} & \multicolumn{1}{c|}{26.85}  & \multicolumn{1}{c||}{0.7420} & \multicolumn{1}{c|}{25.39} & \multicolumn{1}{c|}{0.7558} & \multicolumn{1}{c|}{24.63} & \multicolumn{1}{c}{0.6996} \\ \hline
Ours-RCAN  & \multicolumn{1}{c|}{\textbf{27.53}} & \multicolumn{1}{c|}{\textbf{0.7970}}& \multicolumn{1}{c|}{\textbf{27.34}} & \multicolumn{1}{c||}{\textbf{0.7592}} & \multicolumn{1}{c|}{\textbf{25.50}} & \multicolumn{1}{c|}{0.7584}  & \multicolumn{1}{c|}{\textbf{24.84}} & \multicolumn{1}{c}{0.7034} \\ \bottomrule
\end{tabular}}
\captionsetup{font=small}
\caption{\textbf{Quantitative comparison of M\"obius transformation results on ODIs.} ${\rm (+Transform)}$ denotes that we first employ a scale-specific SR model for image SR and then conduct image-level M\"obius transformation on the SR image. We report on ODI-SR dataset and SUN360 dataset with up-sampling factors $\times 8$ and $\times 16$. \textbf{Bold} indicates the best results.}
\label{tab:msr}
\end{table*}

\begin{figure*}[t]
    \centering
    \includegraphics[width=1\linewidth]{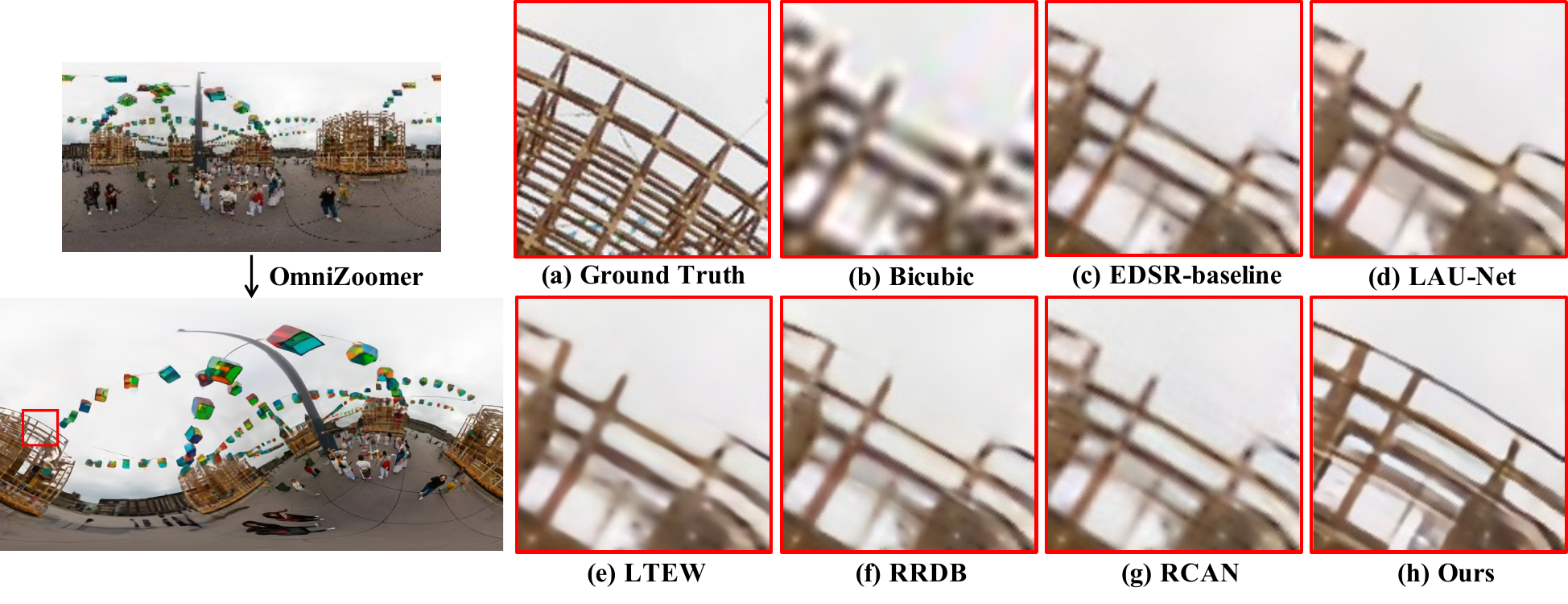}
\captionsetup{font=small}
\caption{Visual comparisons of M\"obius transformation results with $\times 8$ up-sampling factor on ODI-SR dataset.}
\label{fig:comparex8}
\end{figure*}

Inspired by the inherent spherical representation of ODIs and the spherical conformality of M\"obius transformation, we propose the spherical resampling module to generate the transformed feature maps $F_\text{M}$. The spherical resampling module directly resamples on the curved sphere based on the spherical geodesic of two points on the sphere.
Given a query pixel $q$ with the spatial index $(\theta_q, \phi_q)$ from the index map $Y$, we choose its four corner pixels $\{p_i, i=0,1,2,3\}$ as the neighbouring pixels, which are located on the feature maps $F_\text{UP}$ (as shown in the left of Fig.~\ref{fig:7}(b)). 
The indices of the neighboring pixels satisfy the following conditions: $\theta_0=\theta_3$, $\theta_1=\theta_2$, $\phi_0=\phi_1$, and $\phi_2=\phi_3$. To obtain the feature value of the query pixel $q$, we employ the spherical linear interpolation (Slerp)~\cite{Fatelo2021MobilitySA}, 
which is a constant-speed motion along the spherical geodesic of two points on the sphere, formulated as follows:
{\begin{equation}
\small
    \text{Slerp}(a,b) = \frac{\sin(1-t) \beta}{\sin\beta}a + \frac{\sin t \beta}{\sin\beta}b,
    \label{eq:slerp}
\end{equation}}
where $\beta$ is the angle subtended by $a$ and $b$, and $t$ is the resampling weight. Note that $t$ is easy to determine if $a$ and $b$ are located on the same longitude. Therefore, we calculate the feature value of pixel $q$ with two steps. Firstly, we resample $p_0,p_1$ and $p_2,p_3$ to $p_{01}$ and $p_{23}$, respectively, as shown in the right of Fig.\ref{fig:7}(b). Taking the resampling of $p_{0,1}$ as example, the formulation can be described as:
{\begin{equation}
\small
    F(p_{01}) = \frac{\sin(1-t_{01}) \alpha_{01}}{\sin \alpha_{01}}F(p_{0}) + \frac{\sin t_{01} \alpha_{01}}{\sin \alpha_{01}}F(p_{1}),
    \label{eq:6}
\end{equation}}

\noindent where $\alpha_{01}$ is the angle subtended by $p_0$ and $p_1$, and the weight $t_{01}$ is decided by the location of $p_{01}$ on the curve $\overset{\LARGE{\frown}}{p_{0}p_{1}}$. Notably, $t_{01}$ should ensure $p_{01}$ to have the same longitude with the query pixel $q$. Similarly, $\alpha_{23}$ is the angle subtended by $p_2$ and $p_3$, and $p_{23}$ also has the same longitude with the query pixel $q$ by calculating the weight $t_{23}$. After that, we follow the Slerp (Eq.~\ref{eq:slerp}) to calculate the feature value $F_{q}$ as follows:
{\begin{equation}
\small
F(q)=\frac{\sin(1-t_q)\Omega}{\sin\Omega}F(p_{01}) + \frac{\sin t_q\Omega}{\sin\Omega}F(p_{23}),
\end{equation}}

\noindent where $\Omega$ is the angle subtended by $p_{01}$ and $p_{23}$, and $t_q$ is decided by the location of $q$ on the curve $\overset{\LARGE{\frown}}{p_{01}p_{23}}$. \textit{Due to the page limit, more formulations about the parameter $t_{01}$, $t_{23}$ and $t_q$ can be found in the supplementary material.}
Actually, our spherical resampling module calculates the angular relationship of query pixels and their corresponding corner pixels, which can better describe the resampling on the spherical surface with curvatures. Meanwhile, there is no need to estimate the transformed grid shape like~\cite{lee2022learning}, because M\"obius transformation is conformal on the sphere that preserves the angles subtended by two curves.

\section{Experiment}
\label{experiment}

\subsection{Dataset and Implementation Details}
\label{dataset}

\noindent \textbf{Datasets.} No datasets for ODIs under M\"obius transformations exist and collecting real-world ODI pairs with corresponding M\"obius transformation matrices is difficult.  Thus, we propose ODI-M\"obius (ODIM) dataset to train our OmniZoomer and compared methods in a supervised manner. Our dataset is based on the ODI-SR dataset~\cite{deng2022omnidirectional} with 1191 images in the train set, 100 images in the validation set, and 100 images in the test set. During training, we applied various
M\"obius transformations via setting random parameters $\{a,b,c,d\}$ of Eq.~\ref{eq:1}. Each M\"obius transformation includes all horizontal rotation, vertical rotation, and zoom, as we aim to move and zoom in on ODIs. \textit{More details can be found in the Suppl material}. During validating and testing, we assign a fixed M\"obius transformation matrix for each ODI. Besides, we further test on SUN360~\cite{Xiao2012RecognizingSV} dataset with 100 images.

\begin{table*}[t!]
\scalebox{0.91}{
\begin{tabular}{c||cccc||cccc}
\toprule
Scale & \multicolumn{4}{c||}{$\times 8$} & \multicolumn{4}{c}{$\times 16$} \\ \hline
\multirow{2}{*}{Method} & \multicolumn{2}{c|}{ODI-SR} & \multicolumn{2}{c||}{SUN 360} & \multicolumn{2}{c|}{ODI-SR} & \multicolumn{2}{c}{SUN 360} \\ \cline{2-9} & \multicolumn{1}{c|}{WS-PSNR} & \multicolumn{1}{c|}{WS-SSIM} & \multicolumn{1}{c|}{WS-PSNR} & WS-SSIM & \multicolumn{1}{c|}{WS-PSNR} & \multicolumn{1}{c|}{WS-SSIM} & \multicolumn{1}{c|}{WS-PSNR} & WS-SSIM \\ \hline \hline
Bicubic  & \multicolumn{1}{c|}{19.64} & \multicolumn{1}{c|}{0.5908} & \multicolumn{1}{c|}{19.72} & \multicolumn{1}{c||}{0.5403} & \multicolumn{1}{c|}{17.12} & \multicolumn{1}{c|}{0.4332} & \multicolumn{1}{c|}{17.56} & \multicolumn{1}{c}{0.4638} \\ \hline
EDSR~\cite{lim2017enhanced} & \multicolumn{1}{c|}{23.97} & \multicolumn{1}{c|}{0.6483} & \multicolumn{1}{c|}{23.79} & \multicolumn{1}{c||}{0.6472} & \multicolumn{1}{c|}{22.24} & \multicolumn{1}{c|}{0.6090} & \multicolumn{1}{c|}{21.83} & \multicolumn{1}{c}{0.5974} \\ \hline 
RCAN~\cite{zhang2018image} & \multicolumn{1}{c|}{24.26} & \multicolumn{1}{c|}{0.6554} & \multicolumn{1}{c|}{23.88} & \multicolumn{1}{c||}{0.6542}        & \multicolumn{1}{c|}{22.49}   & \multicolumn{1}{c|}{0.6176} & \multicolumn{1}{c|}{21.86}        & \multicolumn{1}{c}{0.5938} \\ \hline
360-SS~\cite{ozcinar2019super} & \multicolumn{1}{c|}{24.14} & \multicolumn{1}{c|}{0.6539} & \multicolumn{1}{c|}{24.19} & \multicolumn{1}{c||}{0.6536} & \multicolumn{1}{c|}{22.35} & \multicolumn{1}{c|}{0.6102} & \multicolumn{1}{c|}{22.10} & \multicolumn{1}{c}{0.5947} \\ \hline
SphereSR~\cite{yoon2022spheresr} & \multicolumn{1}{c|}{24.37} & \multicolumn{1}{c|}{0.6777} & \multicolumn{1}{c|}{24.17} & \multicolumn{1}{c||}{0.6820} & \multicolumn{1}{c|}{22.51} & \multicolumn{1}{c|}{0.6370} & \multicolumn{1}{c|}{21.95} & \multicolumn{1}{c}{0.6342} \\ \hline
LAU-Net~\cite{deng2021lau} & \multicolumn{1}{c|}{24.36} & \multicolumn{1}{c|}{0.6602} & \multicolumn{1}{c|}{24.24} & \multicolumn{1}{c||}{0.6708}  & \multicolumn{1}{c|}{22.52} & \multicolumn{1}{c|}{0.6284} & \multicolumn{1}{c|}{22.05} & \multicolumn{1}{c}{0.6058} \\ \hline 
LAU-Net+~\cite{deng2022omnidirectional} & \multicolumn{1}{c|}{\textbf{24.63}} & \multicolumn{1}{c|}{\textbf{0.6815}} & \multicolumn{1}{c|}{\textcolor{blue}{24.37}} & \multicolumn{1}{c||}{0.6710} & \multicolumn{1}{c|}{\textbf{22.97}} & \multicolumn{1}{c|}{\textbf{0.6316}} & \multicolumn{1}{c|}{\textbf{22.22}} & \multicolumn{1}{c}{0.6111} \\ \hline
Ours-EDSR-baseline  & \multicolumn{1}{c|}{24.48} & \multicolumn{1}{c|}{0.6756} & \multicolumn{1}{c|}{24.31} & \multicolumn{1}{c||}{\textcolor{blue}{0.7019}} & \multicolumn{1}{c|}{22.65} & \multicolumn{1}{c|}{0.6304} & \multicolumn{1}{c|}{22.09} & \multicolumn{1}{c}{\textcolor{blue}{0.6449}} \\ \hline
Ours-RCAN  & \multicolumn{1}{c|}{\textcolor{blue}{24.53}} & \multicolumn{1}{c|}{\textcolor{blue}{0.6797}} & \multicolumn{1}{c|}{\textbf{24.41}} & \multicolumn{1}{c||}{\textbf{0.7106}} & \multicolumn{1}{c|}{\textcolor{blue}{22.66}}  & \multicolumn{1}{c|}{\textcolor{blue}{0.6304}} & \multicolumn{1}{c|}{\textcolor{blue}{22.12}} & \multicolumn{1}{c}{\textbf{0.6454}} \\ \bottomrule
\end{tabular}}
\captionsetup{font=small}
\caption{\textbf{Quantitative comparison of ODI SR task.} The numbers are excerpted from ~\cite{deng2022omnidirectional} except for~\cite{yoon2022spheresr}, due to its reported results are obtained by utilizing 800 training images in the ODI-SR dataset. We report $\times 8$, $\times 16$ SR results on the ODI-SR and SUN360 datasets. Bold indicates the best results, and \textcolor{blue}{blue} indicates the second-best results.}
\label{tab:sr}
\end{table*}

\begin{figure*}[t]
    \centering
    \includegraphics[width=1\linewidth]{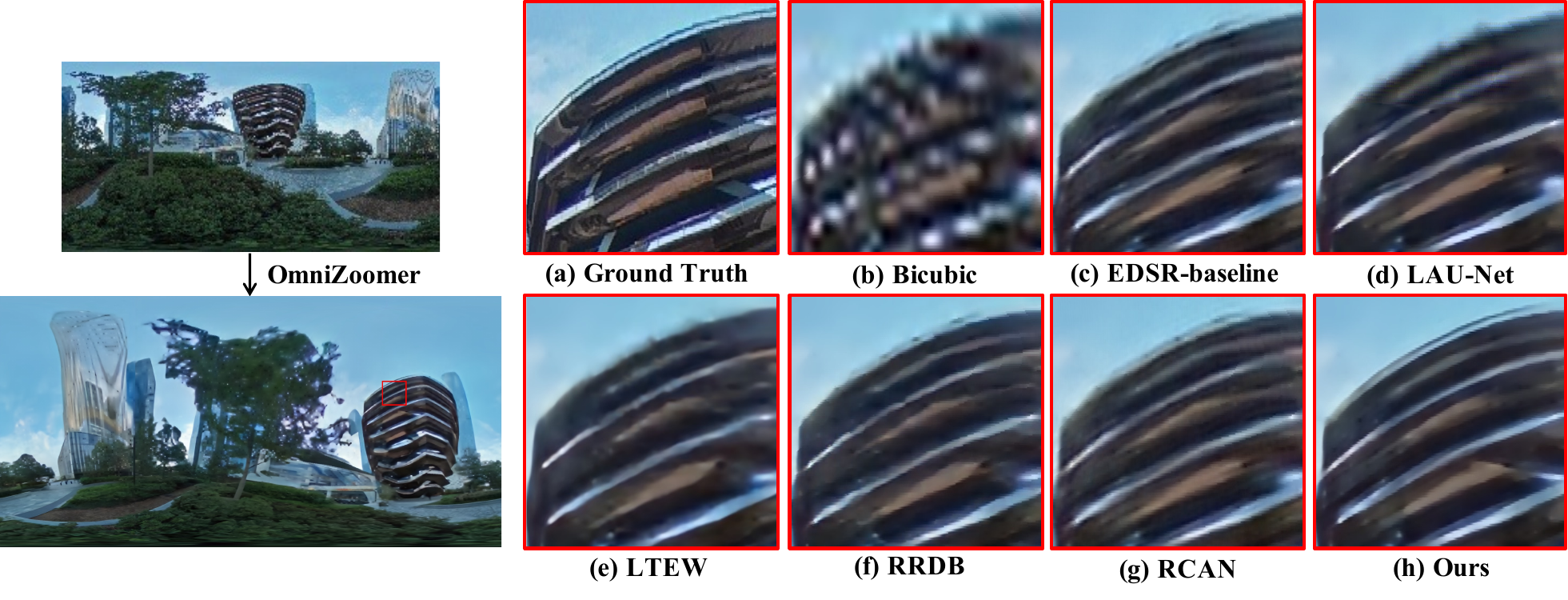}

\captionsetup{font=small}
\caption{Visual comparisons of different methods for M\"obius transformation with $\times 8$ up-sampling factor on SUN360 dataset.}
\label{fig:comparex8_sun}
\end{figure*}

\noindent \textbf{Implementation details.} The resolution of the HR ERP images is $1024 \times 2048$, and the up-sampling factors we choose are $\times8$ and $\times16$. We use L1 loss, which is optimized by
Adam optimizer~\cite{kingma2014adam}, with an initial learning rate of 1e-4. The batch size is 2 when using EDSR-baseline~\cite{lim2017enhanced} as backbone, while the batch size is 1 when using RCAN~\cite{zhang2018image} as backbone. Especially, considering the spherical imagery of ODIs, we use specific WS-PSNR~\cite{sun2017weighted} and WS-SSIM~\cite{zhou2018weighted} metrics for evaluation.

\subsection{Quantitative and Qualitative Evaluation}
\label{comparison}

\noindent \textbf{Move and Zoom in:}
As OminiZoomer is the first learning-based method, there are no prior-arts that can be directly compared. For fair and sufficient evaluation, we design two types of comparative experiment. First, we combine the existing image SR models for 2D planar images and ODIs~\cite{lim2017enhanced,wang2018esrgan,zhang2018image,chao2023equivalent,wang2023omni,deng2021lau} with image-level M\"obius transformations, whose resampling process is achieved by nearest interpolation. This way, we compare OmniZoomer with these approaches under various M\"obius transformations. The SR models designed for 2D planar images are retrained with their provided hyperparameters. Secondly, we further compare our OmniZoomer with existing image warping methods~\cite{son2021srwarp,lee2022learning} and incorporate M\"obius transformations into their learning process.

Tab.~\ref{tab:msr} provides a quantitative comparison of different methods for various M\"obius transformations with up-sampling factors $\times 8$ and $\times 16$. We use a lightweight backbone EDSR-baseline~\cite{lim2017enhanced} and a deep backbone RCAN~\cite{zhang2018image}. 
\textit{OmniZoomer with EDSR-baseline as backbone outperforms several 2D SR models with image-level transformations}, \eg, EDSR-baseline~\cite{lim2017enhanced}, RRDB~\cite{wang2018esrgan} and RCAN~\cite{zhang2018image}, in all metrics. 
It reveals the effectiveness of our OmniZoomer incorporating M\"obius transformation into the feature representation. Compared with the ODI-specific SR method  LAU-Net~\cite{deng2021lau}, our OmniZoomer also achieves better performance. Note that LAU-Net shows lower performance than SR models designed for 2D planar images, \eg, EDSR-baseline. We ascribe it to that LAU-Net is limited to only consider the ODIs captured with vertically placed $360^{\circ}$ cameras, which have no movement and zoom. By applying deeper backbone RCAN, \textit{OmniZoomer outperforms existing methods in all metrics, all up-sampling factors, and test sets}. For example, compared with LAU-Net~\cite{deng2021lau}, OmniZoomer has a 0.57dB improvement of WS-PSNR on SUN360 dataset with $\times 8$ up-sampling factor. As shown in Fig.~\ref{fig:comparex8}, our OmniZoomer predicts clearer wood strips with high-quality textural details with $\times 8$ up-sampling factor, which are missing in other methods' predictions. It shows the effectiveness of our HR feature representation and spherical resampling. Similarly, in Fig.~\ref{fig:comparex8_sun}, OmniZoomer reconstructs more complete structures of the building and preserves the shape of buildings after transformations.

\begin{figure}[t!]
    \centering
    \captionsetup{font=small}
    \includegraphics[width=1\linewidth]{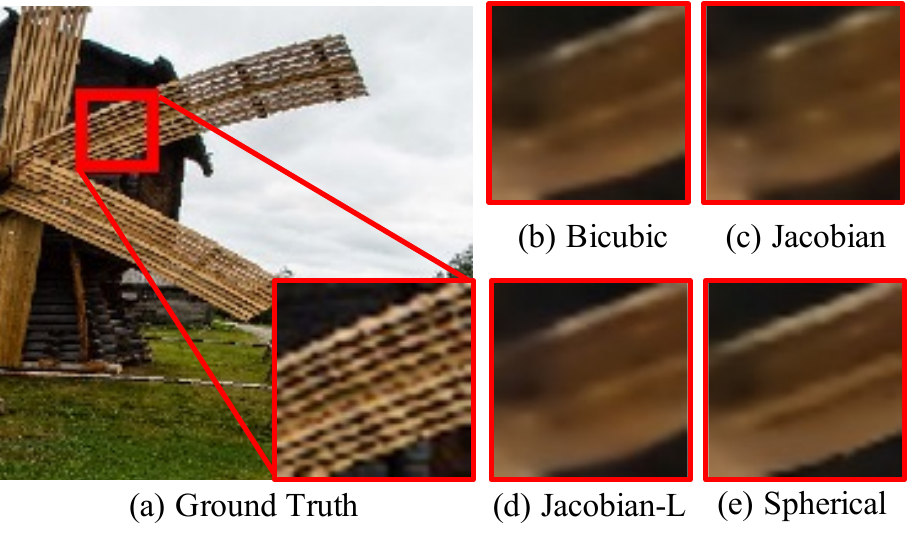}
\caption{Visual comparisons of different resampling methods with $\times 8$ up-sampling factor.}
\label{fig:ab-interpolation}
\end{figure}

\begin{table}[!t]
\setlength{\tabcolsep}{1.3mm}
  \centering
  {
  \begin{tabular}{c | c c}
  \hline
Method & WS-PSNR & WS-SSIM \\
  \hline
 Bicubic & 27.42  & 0.7908 \\
Jacobian~\cite{son2021srwarp} & 27.39 & 0.7909 \\
Jacobian~\cite{son2021srwarp}+MLP & 27.40 & 0.7923 \\
Spherical (Ours) & 27.46 & 0.7930 \\
Spherical+ResBlocks (Ours) & \textbf{27.48} & \textbf{0.7949} \\
  \hline
  \end{tabular}}
  \captionsetup{font=small}
\caption{Ablation results of different \textbf{resampling methods}. We evaluate with $\times 8$ up-sampling factor on ODI-SR dataset.}
\label{table:ab-interpolation}
\end{table}

\begin{table}[!t]
\setlength{\tabcolsep}{0.9mm}
  \centering
  {
  \begin{tabular}{c | c c}
  \hline
Where to apply M\"obius Trans. & WS-PSNR & WS-SSIM \\
  \hline
  Input image level & 26.06  & 0.7621 \\
Input feature level  & 27.03  & 0.7823 \\
HR feature level  & \textbf{27.48}  & \textbf{0.7949} \\
HR Image level & 27.41  & 0.7914 \\
  \hline
  \end{tabular}}
  \captionsetup{font=small}
\caption{Ablation studies on different \textbf{positions for conducting M\"obius transformation}. We evaluate with $\times 8$ up-sampling factor on the ODI-SR dataset.}
\label{table:ab-order}
\end{table}

\noindent \textbf{Direct SR:} Although our work shares a different purpose with the image SR methods, it can also perform SR when the M\"obius transformation matrix is applied as the identity matrix.
In this case, OmniZoomer is degraded to a conventional SR model, except for the spherical resampling module. Tab.~\ref{tab:sr} shows the quantitative results of OmniZoomer with two backbones, exhibiting that OmniZoomer with RCAN as backbone obtains 3 (total 8) best metrics, while OmniZoomer with EDSR-baseline as backbone obtains 2 (total 8) second best metrics. \textit{OmniZoomer has a strong capability to handle the increasing curves and the inherent distortions on ODIs.}

\begin{figure}[t]
\captionsetup{font=small}
    \centering
    \includegraphics[width=.9\linewidth]{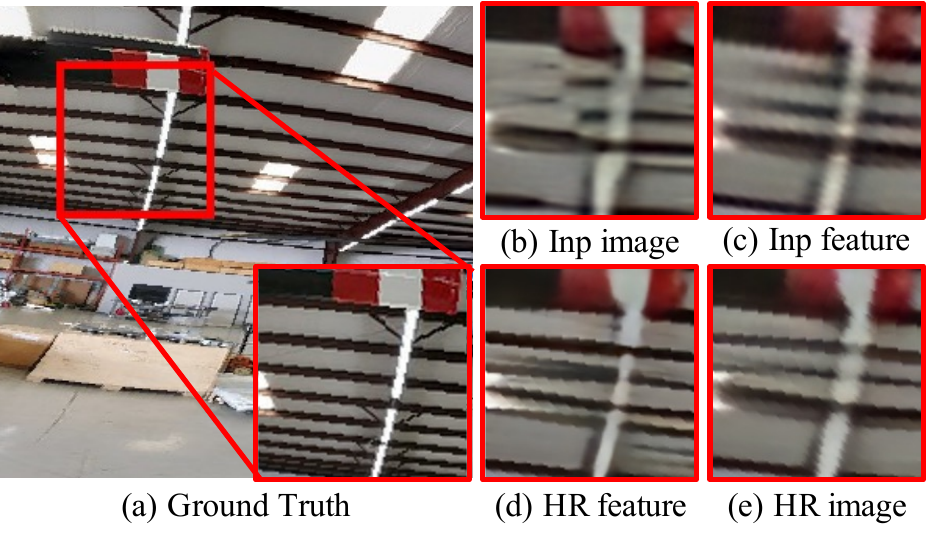}
\caption{Visual comparisons of different positions for conducting M\"obius transformation with $\times 8$ up-sampling factor.}
\label{fig:ab-order}
\end{figure}

\begin{table}[t!]
\setlength{\tabcolsep}{7.5mm}
  \centering
  {
  \begin{tabular}{c | c c}
  \hline
Integration & \ding{51} & \ding{55} \\
  \hline
 WS-PSNR & 27.41 & \textbf{27.48} \\
  \hline
  \end{tabular}}
  \captionsetup{font=small}
\caption{Ablation studies on \textbf{whether to integrate} the two processes of feature up-sampling and M\"obius transformation.}
\label{table:ab-integrate}
\end{table}

\begin{table}[t!]
\setlength{\tabcolsep}{1
mm}
  \centering
  {
  \begin{tabular}{c | c}
  \hline
Methods & Number of parameters \\
  \hline
 LAU-Net~\cite{deng2021lau} & 9.4M \\
 RCAN~\cite{zhang2018image} & 15.9M \\
 OmniZoomer-EDSR-baseline & 1.9M \\
  OmniZoomer-RCAN & 16.0M \\
  \hline
  \end{tabular}}
  \captionsetup{font=small}
\caption{Comparison of the number of parameters (million), which
is conducted on ODI-SR dataset with $\times 8$ up-sampling factor.}
\label{table:params}
\end{table}

\subsection{Ablation Studies}
\label{Ablation study}
\noindent \textbf{Spherical resampling module.} Tab.~\ref{table:ab-interpolation} illustrates that our spherical resampling module achieves the best performance compared with both applying traditional resampling algorithm (\eg, Bicubic), and estimated base rotation in the image warping method~\cite{son2021srwarp}. For example, our spherical resampling module obtains 0.07dB WS-PSNR gain compared with utilizing Jacobian matrix for base rotation. Also, by adding a multi-layer perceptron (MLP) to make the 2D rotation estimation learnable, the performance gain is limited (0.01dB). It is mainly because that~\cite{son2021srwarp} estimates the 2D rotation, which is not applicable for 3D rotation on the sphere surface. Compared with these planar resampling methods, spherical resampling module has obvious improvement benefiting from fitting the sphere surface with curvatures. This can be verified qualitatively in Fig.~\ref{fig:ab-interpolation}, where spherical resampling recovers more continuous edges of the windmill. Furthermore, by adding ResBlocks~\cite{lim2017enhanced} into the decoder, the transformed feature maps can be further refined. It brings 0.02dB gain in WS-PSNR metric.


\noindent \textbf{Different positions for M\"obius transformation.} There are totally four possible positions to conduct M\"obius transformation, \ie, the input image level, the input feature level, the HR feature level, and the HR image level (output of the network). Tab.~\ref{table:ab-order} demonstrates that conducting M\"obius transformation in the input image level and input feature level is not applicable, due to the severely destroyed structures, which are difficult to reconstruct in the HR space. Also, conducting M\"obius transformation on the HR image level leads to sub-optimal results as the network has no knowledge to handle increasing edge curvatures in various transformations, \eg, movement and zoom. From Fig.~\ref{fig:ab-order}, we can see that conducting M\"obius transformation on the HR feature level recovers the clearest pipelines on the ceiling.

\noindent \textbf{Integrating feature up-sampling and M\"obius transformation.} By integrating them into a whole, we find that the performance drops by about 0.07dB WS-PSNR. The reason is about the aliasing problem in the input feature level.

\noindent \textbf{Number of parameters.} Our OmniZoomer with EDSR-baseline as the backbone is compact and performs better than LAU-Net in M\"obius transformation tasks. Also, with 0.1M extra parameters, OmniZoomer-RCAN achieves a significant performance gain than RCAN~\cite{wang2018esrgan}. \textit{More details about computational costs and time consuming of each module can be found in the Suppl material.}

\section{Conclusion}
\label{conclusion}

In this paper, we proposed to incorporate the M\"obius transformation into the network for freely moving and zooming in on ODIs. We found that ODIs under M\"obius transformations suffer from blurry effect and aliasing problems due to zoomed-in regions and increasing edge curvatures. Based on the problems, we found that learning M\"obius transformations on the HR feature level and resampling on the sphere surface enhance the network to predict clear curves and preserved shapes. We demonstrated that this deep learning-based network outperforms existing methods under various M\"obius transformations. Therefore, OmniZoomer can produce HR and high-quality ODIs with the flexibility to move and zoom in to the object of interest.

\noindent \textbf{Limitation and Future Work:} 
This work can estimate HR and high-quality ODIs under various M\"obius transformations. However, the parameters of the M\"obius transformation need to be determined by users, according to the movement and zoom level. In this case, users might try for several times to determine the optimal transformation, which influences the interactive experiences. In the future work, we hope to learn to how to select an optimal transformation by only assigning the interested objects. This might include the techniques about omnidirectional object detection and scene understanding. 

\noindent \textbf{Acknowledgement:} This work was supported by the CCF-Tencent Open
Fund and the National Natural Science Foundation of China
(NSFC) under Grant No. NSFC22FYT45.
{\small
\bibliographystyle{ieee_fullname}
\bibliography{egbib}
}

\end{document}